\documentclass[11pt,a4paper]{article}

\usepackage[margin=1in]{geometry}
\usepackage[T1]{fontenc}
\usepackage[utf8]{inputenc}
\usepackage{lmodern}
\usepackage{microtype}
\usepackage{graphicx}
\usepackage{booktabs}
\usepackage{placeins}
\usepackage{float}
\usepackage{algpseudocode}
\usepackage{amsfonts,amsmath,amssymb}
\usepackage{algorithm}
\usepackage{multirow}
\usepackage[table]{xcolor}
\usepackage{caption}
\usepackage{subcaption}
\usepackage{authblk}
\usepackage[unicode,hidelinks]{hyperref}

\definecolor{gshade}{gray}{0.92}
\title{Observability-Aligned Evidential Masked Pretraining for Reliable VRU Motion Prediction}

\author[1]{Shang Gao\thanks{ORCID: \href{https://orcid.org/0009-0005-5474-1888}{0009-0005-5474-1888}}}

\author[1]{Franz G\"otz-Hahn\thanks{ORCID: \href{https://orcid.org/0000-0003-3465-5040}{0000-0003-3465-5040}}}

\author[2]{Konrad Doll\thanks{ORCID: \href{https://orcid.org/0000-0002-3746-2319}{0000-0002-3746-2319}}}

\author[1]{Bernhard Sick\thanks{ORCID: \href{https://orcid.org/0000-0001-9467-656X}{0000-0001-9467-656X}}}

\affil[1]{Intelligent Embedded Systems, University of Kassel, Germany\\
\texttt{\{sgao,bsick,franz.goetz-hahn\}@uni-kassel.de}}

\affil[2]{Aschaffenburg University of Applied Sciences, Germany\\
\texttt{konrad.doll@th-ab.de}}

\date{}
\hypersetup{
  pdftitle={Observability-Aligned Evidential Masked Pretraining for Reliable VRU Motion Prediction},
  pdfauthor={Shang Gao, Bernhard Sick, Franz Goetz-Hahn}
}

\begin{document}
\maketitle

\begin{abstract}
Vulnerable road user (VRU) prediction must operate under partial observability and distribution shift, so reliable uncertainty matters alongside accuracy. Masked pretraining is attractive in this setting because it naturally exposes the model to missing content during self-supervision, but existing time-series masked autoencoding methods are typically deterministic and do not constrain how confidence should change with observability. We introduce eCoMAE, an evidential masked pretraining method for multivariate time series that explicitly aligns predictive evidence with what is observed. eCoMAE combines block-wise time-step masking with a Normal–Inverse–Gamma decoder and three observability-aware mechanisms: uncertainty-weighted mean consistency on positions visible in both a masked view and a reference pass, evidence suppression on masked positions, and representation consistency across stochastic masks. On IMPTC benchmark for VRU intention prediction under leave-one-environment-out weather shift, eCoMAE improves the strongest baseline from 0.7971 to 0.8068 in OOD accuracy and from 0.0718 to 0.0648 in OOD AURC. Additional evaluation on pedestrian trajectory forecasting across IMPTC, Waymo, nuScenes, and inD further yields the best fine-tuning MSE on all four datasets. These results show that observability-aligned evidential pretraining improves reliability under shift and remains effective beyond the primary benchmark.

\end{abstract}


\vspace{1em}

\begin{figure}[t]
  \centering
  \includegraphics[width=\linewidth]{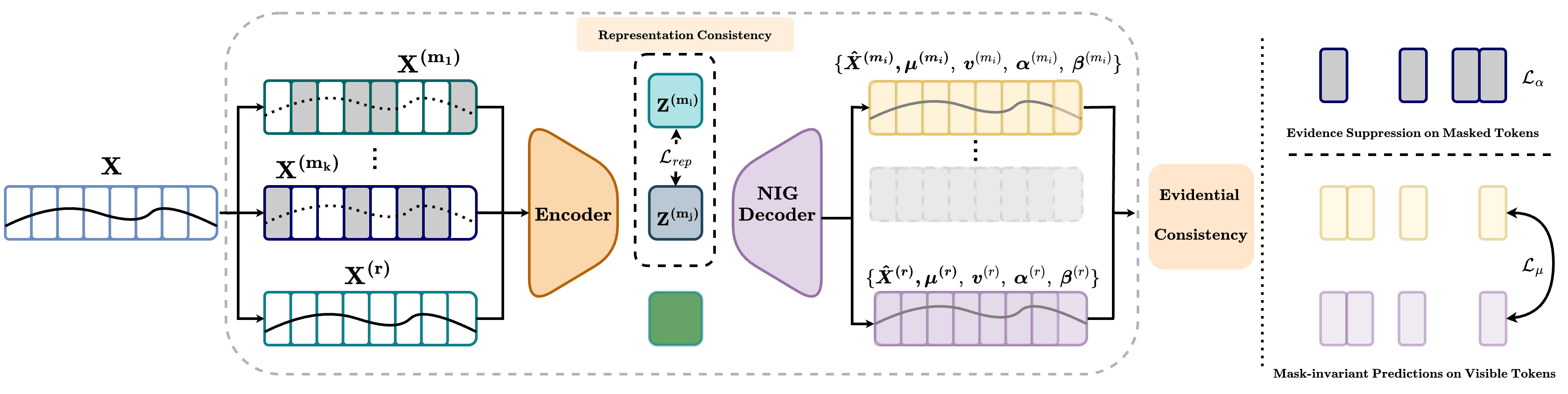}
  \caption{\textbf{eCoMAE workflow.}
    For each input sequence, $K$ stochastic masked views and one unmasked reference view are processed by a shared encoder–decoder with an NIG head. Training combines masked fitting, visible-token mean consistency against the reference, masked-token evidence suppression, and cross-view representation consistency.
    % From an input sequence $\mathbf{X}$, eCoMAE samples $K$ stochastic block-wise time-step masked views $\mathbf{X}^{(m_1)}, \ldots, \mathbf{X}^{(m_K)}$ and one unmasked reference view $\mathbf{X}^{(r)}$ for observability alignment.
    % A shared encoder and an NIG decoder output reconstructions and evidential parameters $(\mu,\nu,\alpha,\beta)$.
    % Training combines masked modelling, representation consistency $\mathcal{L}_{\mathrm{rep}}$ across masked views, evidence suppression on masked tokens $\mathcal{L}_{\alpha}$, and evidential consistency $\mathcal{L}_{\mu}$, where a configurable subset of masked views is aligned to the reference on tokens visible in both views.
    }

  \label{fig:ecomae}
\end{figure} 

\section{Introduction}
Vulnerable road user (VRU) motion prediction is a core capability for intelligent transportation systems and safety-critical automation. It includes both discrete intention recognition and continuous trajectory forecasting, and both settings operate under partial observability, changing sensing conditions, and distribution shift. In such environments, reliable decision making requires not only strong predictive performance but also calibrated uncertainty and confidence rankings that remain meaningful under shift.
% Vulnerable road user (VRU) motion prediction is a core capability for intelligent transportation systems and safety-critical automation. It includes both discrete intention recognition and continuous trajectory forecasting. In both settings, only limited observations of future behaviour are available, sensing quality can vary across conditions, and the training and test environments may differ. Reliable decision making therefore depends not only on strong predictive performance but also on calibrated uncertainty and confidence rankings that remain meaningful under shift.

Masked pretraining is a natural fit for VRU prediction because it exposes the model to missing content during self-supervision. However, existing time-series masked autoencoding methods mainly optimize deterministic reconstruction and leave uncertainty to downstream add-ons such as ensembles~\cite{Lakshminarayanan2017deep_ensembles}, Bayesian approximations~\cite{Gal2016mc_dropout}. As a result, pretraining does not directly constrain how confidence should respond to partial observability.

We instead ask a more specific question: can masked pretraining shape both representation quality and uncertainty formation so that confidence is high on shared visible content and restrained on masked content? eCoMAE answers this by combining an evidential decoder with observability-aligned regularisation. The key novelty is not merely adding evidential outputs to masked autoencoding, but coupling evidence formation to what is actually observed under each mask.

This work proposes \textbf{eCoMAE}, an evidential consistency masked autoencoder for time-series pretraining that learns representations and confidence signals within the self-supervised objective.
As shown in Fig.~\ref{fig:ecomae}, eCoMAE adopts \emph{block-wise time-step token masking} that masks all channels at selected time indices, reducing leakage across features and preserving temporal structure.
A Transformer encoder--decoder architecture is equipped with a Normal-Inverse-Gamma (NIG) evidential head~\cite{Amini2020edl_regression_nig}, producing distributional predictions rather than point reconstructions.
Crucially, eCoMAE aligns confidence with observability through evidential consistency. Predictions on positions that are visible in both a masked view and a reference pass are encouraged to match the reference, while masked positions are discouraged from showing high evidence through an explicit evidence cap. A representation consistency term further stabilises embeddings across different mask realisations. The method supports a configurable compute--performance trade-off: the number of stochastic mask views and the subset used for evidential alignment are configurable, and an optional visible-token encoder reduces pretraining cost at high mask rates. Padding-aware masking and losses also support variable-length sequences with missingness.

eCoMAE is evaluated primarily on four-class VRU intention classification using the public IMPTC~\cite{Hetzel2023imptc} benchmark, under both in-distribution (ID) testing and out-of-distribution (OOD) evaluation induced by a leave-one-environment-out weather split. The evaluation extends beyond predictive performance to include calibration, selective-prediction reliability, extensive sensitivity studies, and targeted ablations of the proposed design. Broader applicability is further assessed by evaluating the method for pedestrian trajectory forecasting on a unified benchmark built from IMPTC, Waymo~\cite{sun2020waymo}, nuScenes~\cite{caesar2020nuscenesmultimodaldatasetautonomous}, and inD~\cite{Bock2020ind}, following Hetzel et al.~\cite{hetzel2024reliable}.

The main contributions are as follows:
\begin{itemize}
    \item \textbf{eCoMAE}, an observability-aligned evidential masked pretraining method for multivariate time series, integrates masked modelling with predictive uncertainty learning.
    % \item \textbf{Observability-aligned confidence learning} is achieved through uncertainty-weighted consistency on overlap-visible positions, explicit evidence suppression on masked positions, and representation consistency across stochastic masking patterns.
    \item \textbf{Observability-aligned confidence learning} is achieved through consistency on overlap-visible positions, evidence suppression on masked positions, and representation consistency across stochastic masks.
    \item A \textbf{comprehensive experimental evaluation} is conducted, centered on IMPTC intention prediction under environment shift, with reliability-focused analysis, sensitivity studies, and targeted ablations, and further supported by additional evaluation on pedestrian trajectory forecasting benchmarks.
\end{itemize}

\section{Related Work}

\paragraph{Self-supervised time series representation learning.}
% Self-supervised representation learning for time series is commonly studied through contrastive and reconstruction based pretext objectives.
% Contrastive learning builds invariances across augmentations to obtain transferable features~\cite{yue2022TS2Vec},  whereas reconstruction-based objectives learn representations by encoding a series and reconstructing the original signal or its temporal context~\cite{Malhotra2017TimeNet}.
% Masked modelling, inspired by masked language modelling and masked autoencoders (MAE)~\cite{He2022mae}, has recently been adapted to temporal signals as a scalable pretraining strategy~\cite{li2023timaeselfsupervisedmaskedtime}.
% Most existing pretext objectives are deterministic and do not learn predictive uncertainty as part of pretraining, which can limit reliability under distribution shift.
Existing time series self-supervised methods are dominated by contrastive and reconstruction-based objectives~\cite{yue2022TS2Vec,Malhotra2017TimeNet,li2023timaeselfsupervisedmaskedtime}. These methods learn transferable features, but their pretraining objectives are typically deterministic and do not shape predictive uncertainty. Reliability under shift is therefore usually addressed only after pretraining.

\paragraph{Uncertainty estimation and evidential learning.}
Uncertainty in deep learning is often estimated downstream using deep ensembles, Bayesian approximations, or post-hoc calibration procedures~\cite{Lakshminarayanan2017deep_ensembles,Gal2016mc_dropout,guo2017calibration}.
Evidential deep learning provides an alternative by predicting distribution parameters directly, enabling uncertainty to be learned end-to-end from the training objective~\cite{Sensoy2018edl_classification,Amini2020edl_regression_nig}.
Related to uncertainty-aware reconstruction, prior work has explored an economical evidential autoencoder (EAE) design that produces distributional outputs from an autoencoding objective with reduced overhead~\cite{Gao2025FEAE}.
However, this line of work is primarily developed for uncertainty-aware autoencoding in a specific downstream setting, rather than for learning broadly transferable time-series representations with predictive uncertainty as part of self-supervised pretraining.
Moreover, directly optimizing distributional losses does not by itself guarantee that confidence remains aligned with partial observability under masking.

\paragraph{Consistency across masked views.}
Consistency across views or perturbations is a central principle in self-supervised learning, including representation alignment and self-distillation approaches~\cite{grill2020bootstraplatentnewapproach,baevski2022data2vecgeneralframeworkselfsupervised}.
In masked pretraining, this issue becomes particularly relevant because different stochastic masks can produce inconsistent predictions for the same underlying content.
This problem has been studied in masked image modelling, where EMAE introduces self-consistency regularisation across overlapping masked regions~\cite{Li2023EMAE}.
More broadly, related work has explored uncertainty-aware consistency regularisation, where uncertainty is used to modulate the strength of alignment according to prediction reliability~\cite{Zhou2020UACR}.
Motivated by these ideas, eCoMAE studies consistency across masked views for multivariate time series and incorporates evidential uncertainty into this consistency regularisation.

\section{Approach}
\label{sec:approach}
This section defines eCoMAE, a masked pretraining objective for multivariate time series that jointly models reconstruction and predictive evidence under partial observability.

\subsection{Overview}
Given a multivariate time-series window $\mathbf{X}\in\mathbb{R}^{L\times F}$, eCoMAE pretrains a Transformer encoder $f_\theta$
through masked modelling at the \emph{time-step token} level.
A masked set of time indices $\mathcal{M}\subset\{1,\dots,L\}$ is sampled on valid (non-padding) steps to form
$\tilde{\mathbf{X}}=g(\mathbf{X};\mathcal{M})$.
From a masked forward pass, the model predicts both a reconstruction and a NIG predictive distribution.
Beyond fitting masked targets, eCoMAE aligns evidential outputs with \emph{observability} induced by masking:
(i) predictions on visible steps should be invariant to masking elsewhere; (ii) masked steps should not receive unwarranted evidence.
In addition, eCoMAE stabilises learned representations across different mask realisations and implements an efficient
visible-token encoder to reduce the computational cost at high mask rates.
eCoMAE supports $K$ stochastic mask views per sample, where $K$ is configurable to trade computation for robustness.
For each view $k$, a block-wise time-step mask $\mathcal{M}^{(k)}$ yields NIG outputs.
Masked fitting losses are averaged across views. 
Representation consistency is computed over all masked-view pairs, whereas evidential consistency is computed only on a selectable subset of masked views against the reference pass.

\subsection{Mask Invariant Predictions on Visible Tokens}
\label{subsec:evmu}
Let $\pi\in\{0,1\}^{L}$ be a padding mask where $\pi_t=1$ denotes padding at time step $t$, and let $v_t=\mathbb{I}[\pi_t=0]$
indicate validity. A masked index set $\mathcal{M}$ is sampled only from valid steps; a block length $b$ is used to encourage
\emph{blocky} masks. The masking operator replaces entire time-step tokens,
\begin{equation}
\tilde{\mathbf{x}}_t=
\begin{cases}
\mathbf{m}, & t\in\mathcal{M},\\
\mathbf{x}_t, & t\notin\mathcal{M},
\end{cases}
\qquad
\tilde{\mathbf{X}}=g(\mathbf{X};\mathcal{M}),
\label{eq:masking}
\end{equation}
where $\mathbf{m}\in\mathbb{R}^{F}$ is a learnable mask embedding. This time-step token masking avoids feature leakage by
masking all $F$ channels at selected steps.

The masked forward pass uses a visible-valid set
\begin{equation}
\mathcal{V}=\{t:\ v_t=1,\ t\notin\mathcal{M}\},\qquad L_{\mathrm{vis}}=|\mathcal{V}|,
\label{eq:visible_set}
\end{equation}
and encodes only the packed visible sequence,
\begin{equation}
\mathbf{Z}_{\mathrm{vis}}^{(m)} = f_\theta(\tilde{\mathbf{X}}_{\mathcal{V}})\in\mathbb{R}^{L_{\mathrm{vis}}\times D}.
\label{eq:enc_vis}
\end{equation}
A full-length decoder reconstructs all positions by placing learned mask tokens at every step and injecting the encoded
visible tokens at their original indices. The decoder outputs
\begin{equation}
(\hat{\mathbf{X}}^{(m)},\boldsymbol{\mu}^{(m)},\boldsymbol{\nu}^{(m)},\boldsymbol{\alpha}^{(m)},\boldsymbol{\beta}^{(m)})
= d_\phi(\mathbf{Z}_{\mathrm{vis}}^{(m)}),
\label{eq:decode}
\end{equation}
where each output lies in $\mathbb{R}^{L\times F}$. Positivity is enforced via standard transforms (e.g., softplus) and
$\alpha>1$ is ensured to obtain a proper NIG predictive distribution.

Define masked-valid and visible-valid index sets
\begin{align}
\Omega_m &= \{(t,f)\mid t\in\mathcal{M},\ v_t=1,\ f\in\{1,\dots,F\}\},\\
\Omega_u &= \{(t,f)\mid t\notin\mathcal{M},\ v_t=1,\ f\in\{1,\dots,F\}\},
\end{align}
and the averaging operator
\begin{equation}
\langle a\rangle_{\Omega} \;=\; \frac{1}{|\Omega|}\sum_{(t,f)\in\Omega} a_{t,f}.
\label{eq:avg_op}
\end{equation}
Masked modelling fits only $\Omega_m$ to prevent leakage:

\begin{equation}
\mathcal{L}_{\mathrm{evi}}
=
\Big\langle
\begin{aligned}
\ell_{\mathrm{NIG}}\!\big(x_{t,f};\mu^{(m)}_{t,f},\nu^{(m)}_{t,f},\alpha^{(m)}_{t,f},\beta^{(m)}_{t,f}\big)
\\
+\ \lambda_{\mathrm{reg}}\,
\ell_{\mathrm{reg}}\!\big(x_{t,f},\mu^{(m)}_{t,f},\nu^{(m)}_{t,f},\alpha^{(m)}_{t,f}\big)
\end{aligned}
\Big\rangle_{\Omega_m}.
\label{eq:l_evi}
\end{equation}

To align predictions with observability, an unmasked \emph{reference pass} is computed on valid steps,
\begin{equation}
(\hat{\mathbf{X}}^{(r)},\boldsymbol{\mu}^{(r)},\boldsymbol{\nu}^{(r)},\boldsymbol{\alpha}^{(r)},\boldsymbol{\beta}^{(r)})
= d_\phi\!\big(f_\theta(\mathbf{X}_{\{t:\,v_t=1\}})\big).
\label{eq:ref_pass}
\end{equation}
EvMu enforces that visible-valid predictions remain invariant to masking elsewhere:
\begin{equation}
\mathcal{L}_{\mu}
=
\Big\langle
\omega_{t,f}\;\phi\!\big(\mu^{(m)}_{t,f}-\mu^{(r)}_{t,f}\big)
\Big\rangle_{\Omega_u},
\qquad
\omega_{t,f}=\frac{1}{\sigma^{2(r)}_{t,f}+\tau},
\label{eq:l_mu}
\end{equation}
where $\phi(\cdot)$ is robust Huber-style penalty, $\tau>0$ is a stabiliser, and
\begin{equation}
\sigma^{2(r)}_{t,f}
=
\frac{\beta^{(r)}_{t,f}\bigl(1+\nu^{(r)}_{t,f}\bigr)}
{\nu^{(r)}_{t,f}\bigl(\alpha^{(r)}_{t,f}-1\bigr)},
\qquad \alpha^{(r)}_{t,f}>1,
\label{eq:nig_pred_var}
\end{equation}
is the predictive variance implied by the reference NIG parameters. This quantity corresponds to the total predictive uncertainty and can be decomposed into an aleatoric term $\beta^{(r)}_{t,f}/(\alpha^{(r)}_{t,f}-1)$ and an epistemic term $\beta^{(r)}_{t,f}/(\nu^{(r)}_{t,f}(\alpha^{(r)}_{t,f}-1))$. The weighting therefore down-weights entries that are intrinsically uncertain under the reference pass, focusing the invariance constraint on stable, well-observed predictions.

\subsection{Evidence Suppression on Masked Tokens}
\label{subsec:evalpha}
Masked supervision alone does not prevent the model from producing overly confident evidence at unobserved locations.
EvAlpha explicitly discourages excessive evidence on masked-valid entries by penalising the evidential strength
\begin{equation}
\Phi^{(m)}_{t,f} = 2\nu^{(m)}_{t,f} + \alpha^{(m)}_{t,f}
\end{equation}
above a cap $\bar{\Phi}$:
\begin{equation}
\mathcal{L}_{\alpha}
=
\Big\langle [\,\Phi^{(m)}_{t,f}-\bar{\Phi}\,]_+ \Big\rangle_{\Omega_m},
\label{eq:l_alpha}
\end{equation}
where $[x]_+=\max(0,x)$. In our implementation, $\bar{\Phi}$ is a fixed hyperparameter that bounds evidential strength on masked positions throughout training.

In practice, the consistency terms need not use all masked views: the EvMu and EvAlpha penalties are computed on a configurable subset of mask views and averaged across the selected views, enabling cost-aware training when $K$ is large.

\subsection{Mask-Consistent Representations and Visible-Token Efficiency}
\label{subsec:rep_eff}
To stabilise representations across missingness patterns, $K$ mask passes are computed.
Representation consistency is applied to all view pairs $(i,j)$ by pooling only \emph{overlap-visible} positions and averaging the cosine penalty over pairs.
\begin{equation}
\mathcal{L}_{\mathrm{rep}}
=
\frac{2}{K(K-1)}\sum_{1\le i<j\le K}
\Big(1-\cos\!\big(\mathbf{z}^{m_i},\mathbf{z}^{m_j}\big)\Big),
\qquad
\mathbf{z}^{m_i}=h_w\!\big(\mathrm{pool}(\mathbf{Z}_{vis}^{m_i})\big),
\end{equation}

We additionally include a standard masked reconstruction term on the predictive mean,
\[
\mathcal{L}_{\mathrm{rec}}=\left\langle \left(\hat{x}^{(m)}_{t,f}-x_{t,f}\right)^2 \right\rangle_{\Omega_m},
\]
averaged across masked views.

The overall pretraining objective combines masked fitting with observability- and mask-consistency regularisation:
\begin{align}
\mathcal{L}_{\mathrm{ssl}}
&=\underbrace{\lambda_{\mathrm{rec}}\mathcal{L}_{\mathrm{rec}}
+\lambda_{\mathrm{evi}}\mathcal{L}_{\mathrm{evi}}}_{\mathcal{L}_{\mathrm{fit}}}
+\underbrace{\lambda_{\mu}\mathcal{L}_{\mu}
+\lambda_{\alpha}\mathcal{L}_{\alpha}
+\lambda_{\mathrm{rep}}\mathcal{L}_{\mathrm{rep}}}_{\mathcal{L}_{\mathrm{cons-reg}}}.
\end{align}

Finally, the visible-token encoder provides an efficiency benefit. Self-attention cost in the encoder reduces from
$\mathcal{O}(L^2)$ to approximately $\mathcal{O}(L_{\mathrm{vis}}^2)$, while the decoder still reconstructs the full
length-$L$ sequence and losses are computed on $\Omega_m$ and $\Omega_u$. At high mask rates, where $L_{\mathrm{vis}}\ll L$,
this yields substantial compute savings without changing the supervision signal or the observability constraints.

% =========================
% Experiments
% =========================
\section{Experiments}
\label{sec:exp}
The experiments are designed to answer three questions: (i) whether eCoMAE improves predictive performance and reliability on the primary IMPTC intention benchmark under environment shift, (ii) which design choices are most responsible for the gains, and (iii) whether the method remains effective beyond the primary setting on different downstream tasks and datasets.

\subsection{Primary Benchmark: IMPTC Intention Prediction}
\label{subsec:imptc}

The primary evaluation is conducted on a trajectory-window benchmark derived from the public IMPTC VRU dataset. Each sample is represented as a fixed-length multivariate time-series window $X \in \mathbb{R}^{L \times F}$. The downstream task is four-class intention prediction, with labels corresponding to straight, left, right, and stop. To mitigate class imbalance, supervised objectives use class-weighted cross-entropy.

Evaluation is performed under both ID and OOD settings. The dataset provides weather-related metadata, which is used to partition samples into discrete environment groups. OOD generalisation is assessed with a leave-one-environment-out protocol, where one environment group is held out for testing and the remaining groups are used for training. ID evaluation uses the standard test split, which follows the same environment mixture as the training data. This setup makes it possible to assess not only predictive performance in the nominal setting, but also robustness and reliability under environment shift.

\subsection{Training Protocol and Evaluation Metrics}
\label{subsec:protocol}

Unless stated otherwise, evaluation follows a three-stage protocol consisting of self-supervised pretraining, linear probing, and end-to-end finetuning. The main analysis focuses on finetuned models, while probing results are reported to assess representation quality before downstream adaptation.

Performance is evaluated with both predictive and uncertainty-aware metrics. Classification quality is measured by accuracy. Calibration is quantified by expected calibration error (ECE) \cite{guo2017calibration}. Under distribution shift, selective-prediction reliability is assessed by the area under the risk--coverage curve (AURC), where lower values indicate better abstention behaviour \cite{geifman2018selective}. Alignment between uncertainty and errors is further measured by uncertainty--error correlation (UE-corr), computed as the Pearson correlation between predictive entropy and the corresponding 0--1 error indicator, where higher values indicate better error awareness \cite{ovadia2019can,kendall2017uncertainties}.

The analysis on IMPTC is organized in two stages. First, a set of model studies examines the role of mask rate, model capacity, evidential sensitivity, and individual objective components. Second, the resulting configuration is compared against supervised, reconstruction-based pretraining, and contrastive self-supervised baselines under both ID and OOD evaluation.

\begin{figure}[t]
  \centering
  \includegraphics[width=0.55\linewidth]{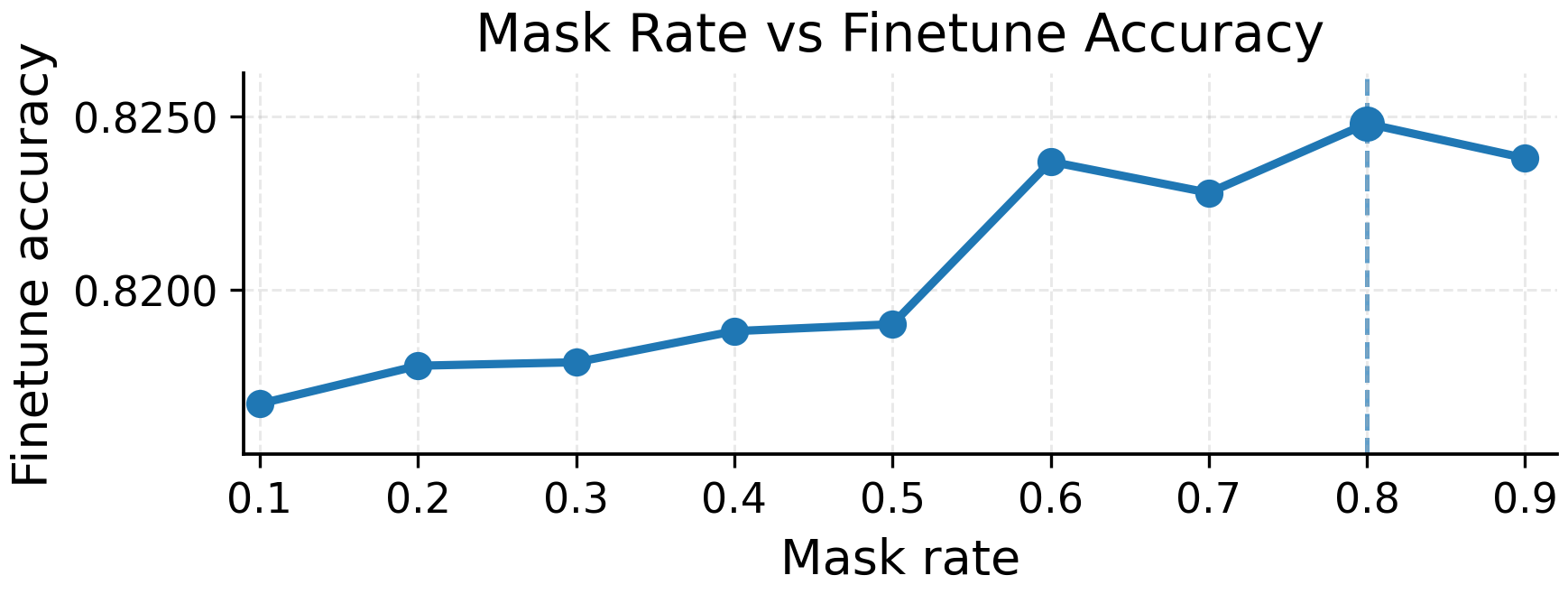}
  \caption{ Mask-rate sensitivity of finetuning accuracy}

  \label{fig:maskrate}
\end{figure} 

\subsection{Model Analysis on IMPTC}
\label{subsec:analysis}
This subsection analyzes the main design choices of eCoMAE on the primary IMPTC benchmark. The study first examines mask-rate selection, then evaluates model capacity, and finally investigates evidential sensitivity and component ablations to isolate the contributions of the proposed objectives.

\paragraph{Mask-rate selection.}
Figure~\ref{fig:maskrate} summarises the mask-rate study. Finetune accuracy improves steadily as the mask rate increases from low to moderate values and reaches its best point around $0.80$, after which the gain saturates. This supports the intended trade-off of masked pretraining: stronger masking increases the difficulty of reconstruction and encourages more informative representations, but overly aggressive masking offers little further benefit. Based on the joint consideration of accuracy, calibration, and computational cost, the default mask rate is set to $0.80$.
\begin{table}[t]
\centering
\scriptsize
\setlength{\tabcolsep}{5pt}
\renewcommand{\arraystretch}{1.25}

\newcommand{\grouprowSix}[1]{%
  \rowcolor{gshade}\multicolumn{6}{@{}l@{}}{\textbf{#1}}\\[-1pt]
}

\begin{tabular}{@{}lccccc@{}}
\toprule
\multirow{2}{*}{\textbf{Capacity}} &
\multicolumn{2}{c}{\textbf{ID finetune}} &
\multicolumn{2}{c}{\textbf{OOD finetune}} &
\textbf{Cost} \\
\cmidrule(lr){2-3}\cmidrule(lr){4-5}\cmidrule(lr){6-6}
& Acc$_f\uparrow$ & ECE$_f\downarrow$
& Acc$_f\uparrow$ & ECE$_f\downarrow$
& min$\downarrow$ \\
\midrule

\grouprowSix{Capacity scaling at MR=0.80}
Small   & 0.8040 & 0.0384 & 0.7824 & \textbf{0.0642} & \textbf{40.5} \\
Medium  & 0.8247 & 0.0184 & 0.7941 & 0.0781          & 41.4 \\
Large   & \textbf{0.8384} & \textbf{0.0129} & \textbf{0.7971} & 0.0698 & 48.7 \\
\bottomrule
\end{tabular}

\caption{Capacity sweep results and backbone definitions. Small, Medium, and Large set $(d_{model}, d_{ff}, E, D, H)$ to (32, 64, 2, 1, 4), (64, 128, 2, 1, 8), and (128, 256, 3, 2, 8), respectively. Here $d_{model}$ is the Transformer hidden size, $d_{ff}$ is the width of the feed-forward network inside each Transformer block, $E$ is the number of encoder layers, $D$ is the number of decoder layers, and $H$ is the number of attention heads.}
\label{tab:capacity_sweep_finetune_cost}
\end{table}
\paragraph{Capacity analysis.}
Table~\ref{tab:capacity_sweep_finetune_cost} reports the capacity sweep at the selected mask rate. Increasing capacity improves finetune accuracy in both ID and OOD settings, with the large model achieving the best ID accuracy, the best ID calibration, and the best OOD accuracy. The small model is cheapest and attains the best OOD ECE, but at a clear accuracy cost. The medium model offers a similar runtime to the small model while substantially narrowing the performance gap. Since the large model provides the strongest overall predictive performance with only a moderate increase in cost, it is used as the default setting for the subsequent ablation studies unless noted otherwise.

\begin{table*}[t]
\centering
\scriptsize
\setlength{\tabcolsep}{4.5pt}
\renewcommand{\arraystretch}{1.18}

% --- group row helper (9 columns in this table) ---
\newcommand{\grouprow}[1]{%
  \rowcolor{gshade}\multicolumn{9}{@{}l@{}}{\textbf{#1}}\\[-1pt]
}

\begin{tabular}{@{}lcccccccc@{}}
\toprule
\midrule
\multirow{2}{*}{\textbf{Setting}} &
\multicolumn{4}{c}{ID} &
\multicolumn{4}{c}{OOD} \\
\cmidrule(lr){2-5}\cmidrule(lr){6-9}
& Acc$_p$ & ECE$_p$ & Acc$_f$ & ECE$_f$
& Acc$_p$ & ECE$_p$ & Acc$_f$ & ECE$_f$ \\
\midrule

\grouprow{Evidential objective (REP off)}
MAE (no EVI/REP)     & 0.7987 & 0.0436 & 0.8379 & 0.0226 & 0.7794 & 0.0721 & 0.8000 & 0.0799 \\
EVI (no REP)       & 0.7993 & 0.0434 & 0.8387 & 0.0124 & 0.7765 & 0.0595 & 0.8000 & 0.0658 \\
\midrule
\addlinespace[2pt]

\grouprow{Representation consistency (EVI on)}
$\lambda_{\mathrm{rep}}=0.05$         & 0.7998 & 0.0436 & 0.8394 & 0.0148 & 0.7765 & 0.0557 & 0.7971 & 0.0660 \\
eCoMAE ($\lambda_{\mathrm{rep}}=0.10$)         & 0.7996 & 0.0436 & 0.8385 & \textbf{0.0111} & 0.7765 & 0.0563 & \textbf{0.8029} & 0.0681 \\
$\lambda_{\mathrm{rep}}=0.20$         & 0.7995 & 0.0442 & 0.8387 & 0.0137 & 0.7765 & 0.0570 & \textbf{0.8029} & 0.0698 \\
\midrule
\addlinespace[2pt]

\grouprow{Evidential consistency (EVI + REP on)}
No $\mu/\alpha$ consistency         & 0.7999 & 0.0375 & 0.8380 & 0.0130 & 0.7765 & 0.0566 & \textbf{0.8029} & 0.0738 \\
$\mu$ consistency only               & 0.7998 & 0.0425 & 0.8386 & 0.0127 & 0.7765 & 0.0555 & \textbf{0.8029} & 0.0728 \\
$\alpha$ consistency only            & 0.7994 & 0.0425 & 0.8387 & 0.0123 & 0.7765 & 0.0555 & 0.8000 & 0.0631 \\
2$\times$ $\mu/\alpha$ weights                        & 0.7994 & 0.0429 & \textbf{0.8397} & 0.0137 & 0.7765 & 0.0582 & 0.8000 & 0.0658 \\
\midrule
\addlinespace[2pt]

\grouprow{Evidential loss weight ($\lambda_{\mathrm{evi}}$)}
$\lambda_{\mathrm{evi}}=0.1$   & 0.8004 & 0.0430 & 0.8383 & 0.0128 & 0.7765 & 0.0576 & 0.7971 & 0.0620 \\
$\lambda_{\mathrm{evi}}=5$  & 0.7974 & 0.0332 & \textbf{0.8397} & 0.0118 & 0.7706 & 0.0433 & 0.7971 & 0.0624 \\
\midrule
\addlinespace[2pt]

\grouprow{Regularisation and evidence cap}
$\lambda_{\mathrm{reg}}=0$            & 0.7994 & 0.0453 & 0.8395 & 0.0121 & 0.7765 & 0.0542 & \textbf{0.8029} & 0.0594 \\
$\lambda_{\mathrm{reg}}=10^{-3}$        & 0.7993 & 0.0453 & 0.8385 & 0.0142 & 0.7765 & 0.0617 & 0.8000 & \textbf{0.0584} \\
$\lambda_{\mathrm{reg}}=0.1$          & 0.7990 & 0.0386 & 0.8386 & 0.0129 & 0.7765 & 0.0575 & 0.7971 & 0.0749 \\
$\bar{\Phi}=2.5$                  & 0.7994 & 0.0425 & 0.8385 & 0.0119 & 0.7765 & 0.0582 & 0.8000 & 0.0704 \\
$\bar{\Phi}=6$                    & 0.7996 & 0.0423 & 0.8395 & 0.0122 & 0.7765 & 0.0586 & \textbf{0.8029} & 0.0695 \\
\midrule
\addlinespace[2pt]

\grouprow{REP-only controls (EVI off)}
$\lambda_{\mathrm{rep}}=0.10$  & 0.7995 & 0.0466 & 0.8395 & 0.0139 & 0.7765 & 0.0953 & 0.8000 & 0.0679 \\
$\lambda_{\mathrm{rep}}=0.20$  & 0.7995 & 0.0396 & 0.8376 & 0.0137 & 0.7765 & 0.0519 & 0.8000 & 0.0663 \\
\midrule
\bottomrule
\end{tabular}

\caption{Main results of 18 variants under ID and OOD. Subscripts $p$ and $f$ denote probing and finetuning. Higher accuracy and lower ECE are better. Best finetuning values are in bold. In the component ablations, NIG and REP remain enabled; only the $\mu$/$\alpha$ consistency terms vary.}
\label{tab:main_18_variants_compact}
\end{table*}

\begin{figure*}[t]
  \centering
  \includegraphics[width=0.85\linewidth]{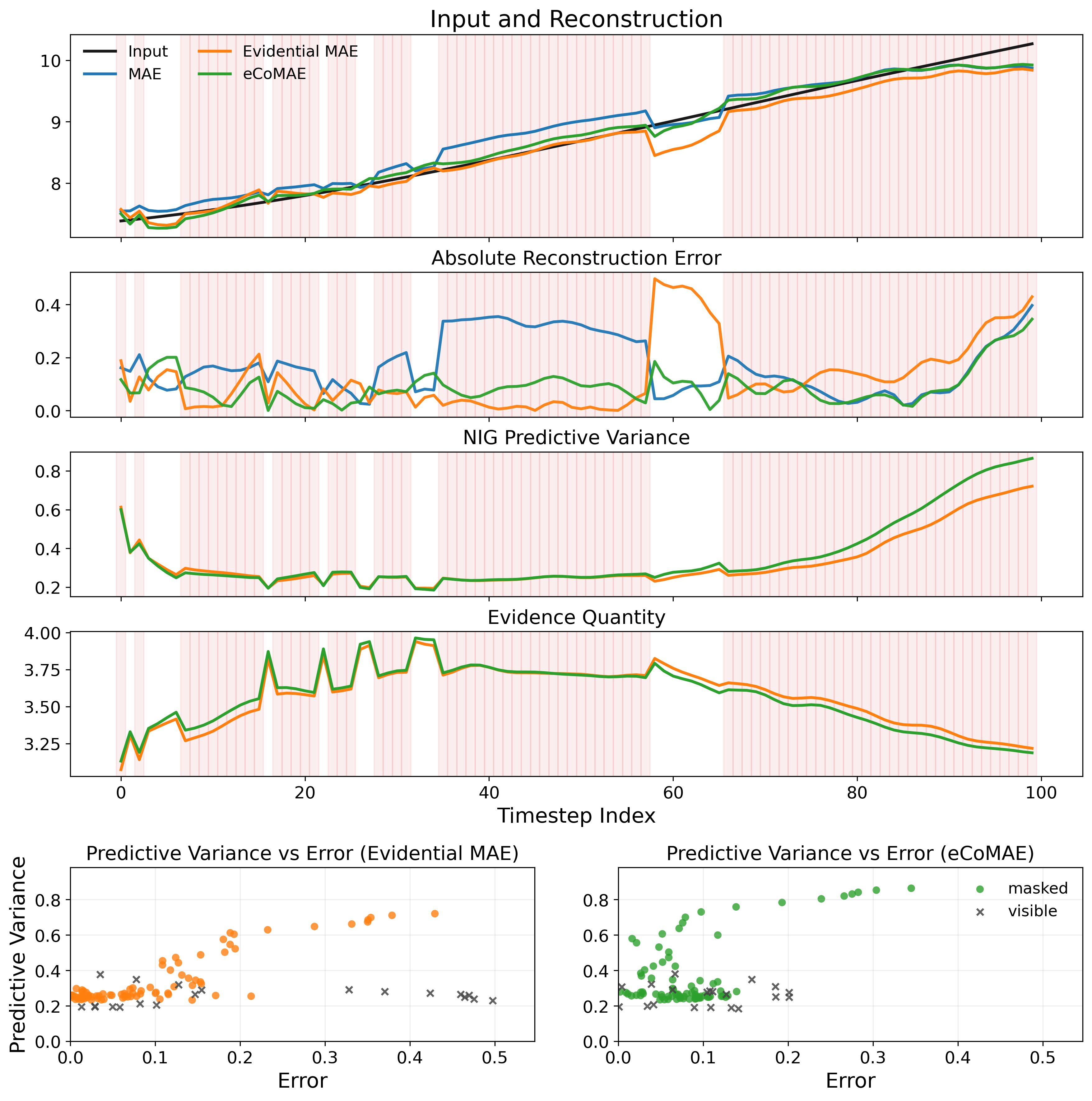}
  \caption{Representative pretraining diagnostic for one sample-feature pair. Shaded regions indicate masked positions. Compared with Evidential MAE, adding representation consistency (eCoMAE) yields lower reconstruction error on masked regions and a clearer increase of predictive variance with reconstruction error, supporting better observability-aligned uncertainty during self-supervised pretraining.}
  \label{fig:evi_pretrain_diag}
\end{figure*}

\paragraph{Evidential sensitivity and component ablations.}
This study analyses how the evidential objective and its components affect robustness and uncertainty under environment shift.
The 18 variants are grouped into six categories:
(i) a core comparison between a plain MAE objective and its evidential counterpart,
(ii) a sweep over mask-wise representation consistency controlled by $\lambda_{\mathrm{rep}}$,
(iii) ablations of the mean and evidence consistency terms with NIG and REP retained,
(iv) sensitivity to the evidential weight $\lambda_{\mathrm{evi}}$,
(v) sensitivity to the evidential regularizer $\lambda_{\mathrm{reg}}$ and the evidence cap $\bar{\Phi}$,
and (vi) sanity checks that enable representation consistency without the evidential objective.
The settings follow the training objective in Sec.~\ref{sec:approach}.
Here, EVI denotes the NIG negative log-likelihood together with evidential regularisation, REP denotes mask-wise representation consistency,
$\lambda_{\mathrm{evi}}$ weights the evidential term,
$\lambda_{\mathrm{reg}}$ penalises excessive evidence when prediction error is large,
$\bar{\Phi}$ caps the evidence-related parameter,
and $\lambda_{\mu}$, $\lambda_{\alpha}$, and $\lambda_{\mathrm{rep}}$ control visible-token mean consistency, masked-token evidence suppression, and representation consistency across mask realisations, respectively. In the component ablations, NIG and REP are kept enabled, so ``only'' refers exclusively to which of the $\mu$ and $\alpha$ consistency terms is active.
\begin{figure*}[t]
  \centering
  \includegraphics[width=0.85\linewidth]{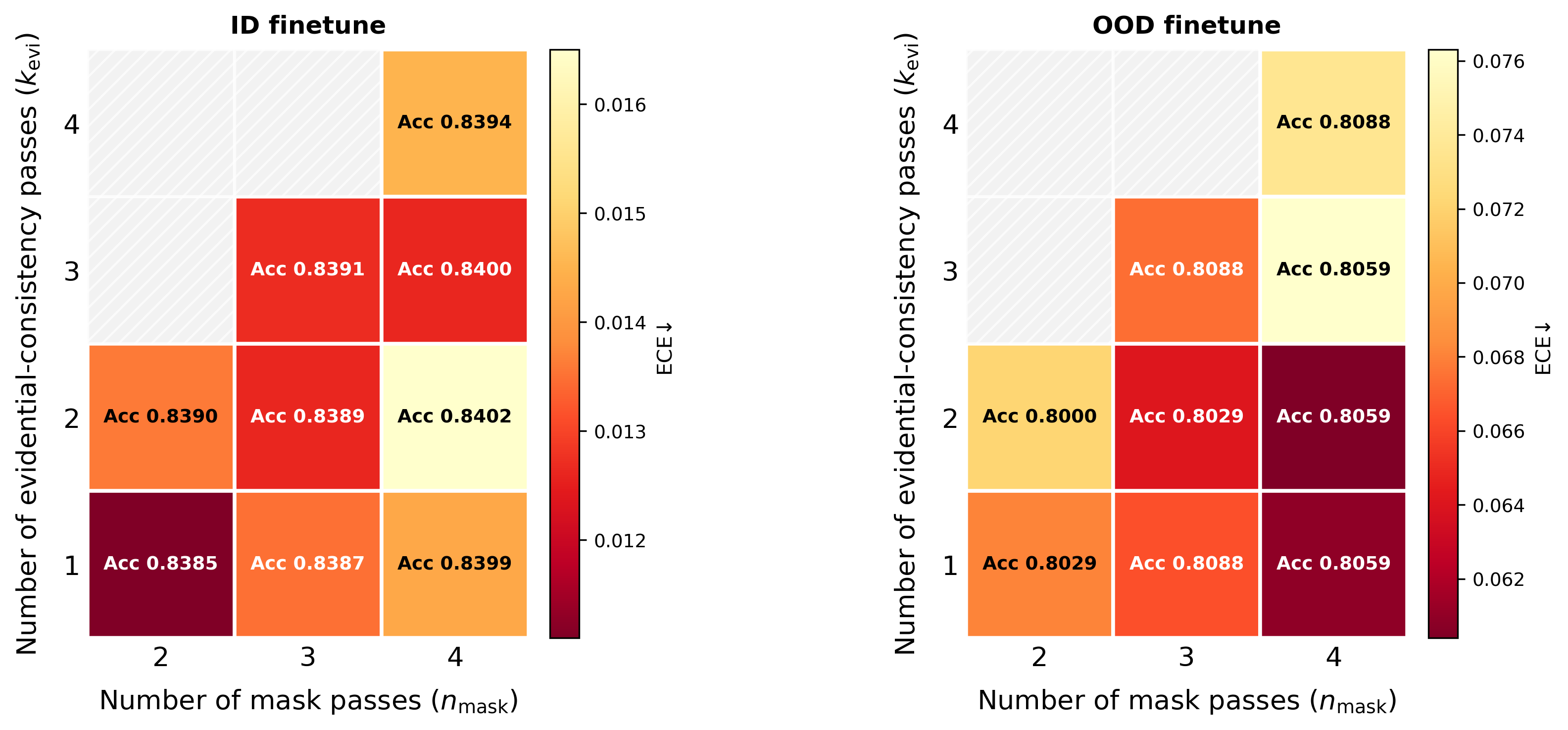}
  \caption{Multi-view scaling design of eCoMAE under different numbers of mask views ($n_{\mathrm{mask}}$) and evidential-consistency views ($k_{\mathrm{evi}}$). Cell text reports finetune accuracy, while color indicates ECE, for both ID and OOD evaluation. Hatched cells denote invalid settings with $k_{\mathrm{evi}} > n_{\mathrm{mask}}$.}
  \label{fig:scalability_kmask_kevi}
\end{figure*}

\begin{table*}[t]
\centering
\scriptsize
\setlength{\tabcolsep}{1.7pt}
\renewcommand{\arraystretch}{1.15}

\begin{tabular}{@{}lcccccccc@{}}
\toprule
\textbf{Method} &
\multicolumn{4}{c}{\textbf{ID}} &
\multicolumn{4}{c}{\textbf{OOD}} \\
\cmidrule(lr){2-5}\cmidrule(lr){6-9}
& Acc$_f\uparrow$ & ECE$_f\downarrow$ & AURC$_f\downarrow$ & UE-corr$_f\uparrow$
& Acc$_f\uparrow$ & ECE$_f\downarrow$ & AURC$_f\downarrow$ & UE-corr$_f\uparrow$ \\
\midrule

\rowcolor{gshade}
\textbf{eCoMAE-S} &
0.8387 & \textbf{0.0123} & 0.0388 & 0.5380 &
0.8014 & 0.0701 & 0.0667 & 0.4464 \\

\rowcolor{gshade}
\textbf{eCoMAE-M} &
0.8389 & 0.0129 & 0.0390 & 0.5388 &
\textbf{0.8068} & \textbf{0.0659} & 0.0672 & 0.4186 \\

\rowcolor{gshade}
\textbf{eCoMAE-L} &
\textbf{0.8397} & 0.0144 & \textbf{0.0382} & \textbf{0.5391} &
0.8066 & 0.0678 & \textbf{0.0648} & 0.4323 \\

\midrule
Transformer (sup.) &
0.7838 & 0.0516 & 0.0991 & 0.4145 &
0.7676 & 0.0825 & 0.1497 & 0.2969 \\

\midrule
Ti-MAE~\cite{li2023timaeselfsupervisedmaskedtime} &
0.8379 & 0.0226 & 0.0409 & 0.5231 &
0.7971 & 0.0857 & 0.0718 & 0.4255 \\

Gaussian Ti-MAE &
0.8367 & 0.0205 & 0.0431 & 0.5212 &
0.7912 & 0.0735 & 0.0748 & \textbf{0.4595} \\

DAE~\cite{vincent2008denoising} &
0.7848 & 0.0145 & 0.0892 & 0.4418 &
0.7706 & 0.0724 & 0.1428 & 0.2877 \\

EAE~\cite{Gao2025FEAE} &
0.7841 & 0.0232 & 0.0934 & 0.4322 &
0.7647 & 0.0665 & 0.1527 & 0.3029 \\

\midrule
SimCLR-TS~\cite{Johannes2022simclr} &
0.7823 & 0.0255 & 0.0738 & 0.5087 &
0.7765 & 0.0815 & 0.1592 & 0.2181 \\

TS2Vec~\cite{yue2022TS2Vec} &
0.7850 & 0.0374 & 0.0933 & 0.4345 &
0.7710 & 0.0869 & 0.1655 & 0.1874 \\

\bottomrule
\end{tabular}
\caption{Main comparison. Higher Acc and UE-corr are better. Lower ECE and AURC are better. Best values are shown in bold. All self-supervised methods are evaluated by end-to-end fine-tuning. DAE denotes a denoising autoencoder baseline. Results are averaged over five random seeds.}
\label{tab:main_compare_ecomae_vs_baselines_joint}
\end{table*}

%%%%%%%%%%%%%%%%%%%%%
\begin{table*}[t]
\centering
\scriptsize
\setlength{\tabcolsep}{3.2pt}
\renewcommand{\arraystretch}{1.15}

\begin{tabular}{@{}lcccc@{}}
\toprule
\textbf{Method} &
\multicolumn{4}{c}{\textbf{Finetune MSE}$\downarrow$} \\
\cmidrule(lr){2-5}
& \textbf{IMPTC} & \textbf{Waymo} & \textbf{nuScenes} & \textbf{inD} \\
\midrule

\rowcolor{gshade}
\textbf{eCoMAE-L} &
\textbf{0.2691} & \textbf{0.2912} & \textbf{0.2514} & \textbf{0.3483} \\

\midrule
Transformer (sup.) &
0.3209 & 0.4809 & 0.3279 & 0.4113 \\

\midrule
Ti-MAE &
0.2997 & 0.3317 & 0.2678 & 0.3753 \\

Gaussian Ti-MAE &
0.2977 & 0.3328 & 0.2679 & 0.3752 \\

DAE &
0.2857 & 0.3012 & 0.3139 & 0.3756 \\

EAE &
0.3571 & 0.2984 & 0.3209 & 0.4628 \\

\midrule
SimCLR-TS &
0.3099 & 0.3015 & 0.2773 & 0.3896 \\

TS2Vec &
0.3108 & 0.2970 & 0.2756 & 0.4528 \\

\bottomrule
\end{tabular}
\caption{Additional evaluation on pedestrian trajectory forecasting on four datasets. Results report finetune MSE under a shared forecasting protocol, lower is better. Best values are shown in bold. Results are averaged over five random seeds.}
\label{tab:compare_ecomae_vs_baselines_joint_traj_forecast}
\end{table*}
Table~\ref{tab:main_18_variants_compact} reports ID and OOD accuracy and calibration for both probing and finetuning. The full hyperparameter dictionary and additional OOD uncertainty diagnostics are provided in Appendix~\ref{app:ablations} (Tables~\ref{tab:evi_config_dictionary_correct} and~\ref{tab:ood_aurc_uecorr_compact}).
In the core comparison, adding EVI leaves OOD finetuning accuracy unchanged at 0.8000 but improves OOD calibration, reducing OOD finetune ECE from 0.0799 to 0.0658; the corresponding OOD probing ECE decreases from 0.0721 to 0.0595 (Table~\ref{tab:main_18_variants_compact}). This indicates that the main effect of the evidential objective is improved reliability under shift rather than improved separability.

Adding REP on top of EVI mainly changes the trade-off between OOD accuracy and uncertainty quality.
Moderate REP ($\lambda_{\mathrm{rep}}=0.10$ or $0.20$) improves OOD finetuning accuracy to 0.8029, whereas $\lambda_{\mathrm{rep}}=0.05$ yields an OOD probe ECE of 0.0557 (Table~\ref{tab:main_18_variants_compact}). This suggests that representation consistency is helpful, but its strength should remain moderate.
The component ablations indicate complementary roles. The $\lambda_{\mu}$-only variant best preserves OOD finetune accuracy among the component-only settings, whereas the $\lambda_{\alpha}$-only variant yields the best OOD finetune ECE. Strengthening all consistency terms mainly benefits ID performance rather than OOD generalisation.

The sensitivity results reveal non-monotonic behaviour.
Changing $\lambda_{\mathrm{evi}}$ away from the default improves some probing uncertainty metrics, but reduces OOD finetuning accuracy.
For $\lambda_{\mathrm{reg}}$, mild settings work best: $\lambda_{\mathrm{reg}}=0$ gives the strongest OOD finetuning accuracy, while $\lambda_{\mathrm{reg}}=10^{-3}$ gives the best OOD finetuning ECE.
By contrast, changing the evidence cap produces only modest differences in this range.
The sanity checks show that REP without EVI is not sufficient to improve OOD accuracy, although it still improves calibration relative to the plain MAE baseline.

Figure~\ref{fig:evi_pretrain_diag} provides a representative diagnostic from the self-supervised pretraining stage and is consistent with the downstream trends. Relative to Evidential MAE, adding Representation Consistency yields lower reconstruction error on masked regions together with a clearer increase of predictive variance as reconstruction error grows, indicating better observability-aligned uncertainty before downstream finetuning.

A related question is how much of the multi-view design is needed in practice. 
Figure~\ref{fig:scalability_kmask_kevi} studies the interaction between the total number of stochastic mask passes $n_{\mathrm{mask}}$ and the number of passes used for evidential consistency $k_{\mathrm{evi}}$. Increasing $n_{\mathrm{mask}}$ generally improves finetune accuracy, especially under OOD evaluation, indicating that additional masked views provide a stronger pretraining signal. At the same time, strong results are already obtained with $k_{\mathrm{evi}} < n_{\mathrm{mask}}$, showing that evidential alignment does not need to be applied exhaustively to every view. This supports a controllable compute--performance trade-off in eCoMAE: more stochastic views are useful, but selective evidential consistency is often sufficient to retain a favorable accuracy--calibration trade-off.

\subsection{Comparison with Baselines on IMPTC}
\label{subsec:main_results}

Table~\ref{tab:main_compare_ecomae_vs_baselines_joint} compares eCoMAE with supervised training, reconstruction-based pretraining, and contrastive self-supervised baselines on the primary IMPTC intention prediction benchmark. Across the eCoMAE family, ID performance is consistently strong, with the large variant achieving the best finetune ID accuracy of $0.8397$ and the best ID AURC of $0.0382$. Under OOD evaluation, the medium variant attains the highest finetune accuracy of $0.8068$, while the large variant achieves the best AURC of $0.0648$. In comparison, the strongest non-eCoMAE baseline, Ti-MAE, reaches $0.7971$ OOD accuracy and $0.0718$ OOD AURC. This shows that eCoMAE improves not only predictive performance under shift but also selective-prediction reliability.

Compared with deterministic masked pretraining and Gaussian or evidential reconstruction variants, eCoMAE more consistently improves OOD reliability. Gaussian Ti-MAE is constructed as a direct Gaussian counterpart to the deterministic Ti-MAE baseline by replacing the point-wise MSE reconstruction loss with a Gaussian negative log-likelihood (NLL) head that predicts a mean and variance. Similarly, EAE replaces the deterministic reconstruction objective with a Normal--Inverse--Gamma negative log-likelihood for masked reconstruction, providing distributional outputs without any observability-aware mechanisms. Together, these comparisons support the central claim that the gains arise from observability-aligned evidence shaping rather than from stronger reconstruction alone.

\subsection{Evaluation on Pedestrian Trajectory Forecasting}
\label{subsec:forecast_transfer}

To assess applicability beyond intention prediction, all methods are further evaluated on pedestrian trajectory forecasting using a unified benchmark constructed from IMPTC, Waymo, nuScenes, and inD, following Hetzel et al.~\cite{hetzel2024reliable}. Under this protocol, each method is first self-supervisedly pretrained on the target dataset and then fine-tuned for forecasting. Table~\ref{tab:compare_ecomae_vs_baselines_joint_traj_forecast} reports the resulting fine-tuning mean squared error on all four datasets. The large eCoMAE variant is shown because it provided the strongest overall pretraining quality in the model-selection stage.

eCoMAE achieves the lowest fine-tuning MSE on all four benchmarks, with values of $0.2691$ on IMPTC, $0.2912$ on Waymo, $0.2514$ on nuScenes, and $0.3483$ on inD. By contrast, the strongest competing baselines show mixed behaviour across datasets and do not match this consistency. These results are best interpreted as evidence that the method remains effective beyond the primary intention-prediction benchmark, rather than as a complete forecasting study.

% =========================
% Discussion
% =========================
\section{Discussion}
\label{sec:discussion}

The results support a more specific interpretation of eCoMAE than simply
``evidential masked autoencoding works better.'' The main empirical pattern is
that the gains concentrate on reliability under shift rather than on large jumps
in separability. In the primary IMPTC evaluation, the strongest improvements
appear in calibration- and selective-risk-related metrics, while accuracy remains
competitive and only moderately improved. This is consistent with the intended
role of the method: the pretraining objective does not merely reconstruct masked
content, but also shapes how confidence should depend on observability. In this
sense, the benefit of eCoMAE is best understood as better \emph{uncertainty
formation} during representation learning.

The ablations also clarify that the three components play different roles rather
than contributing symmetrically. Visible-token mean consistency encourages
agreement where information is actually shared across views, masked-token
evidence suppression discourages overconfident predictions on unobserved content,
and representation consistency stabilises the latent space across stochastic masks.
The model studies suggest that these mechanisms are complementary but must be
balanced carefully. Increasing mask rate or consistency strength can strengthen
the self-supervised signal up to a point, but overly strong constraints reduce
downstream adaptability. Likewise, the evidential sensitivity analysis suggests
that mild regularisation is more effective than either no evidential pressure or
aggressive evidence control. The method therefore works not by maximising all
regularisers simultaneously, but by introducing enough structure to align
confidence with observability while preserving flexibility for fine-tuning.

The additional evaluation results further indicate that the learned representations are not
limited to a single downstream formulation. Although the deepest analysis in
this paper remains centred on intention prediction under environment shift,
fine-tuning on pedestrian trajectory forecasting across IMPTC, Waymo, nuScenes,
and inD shows that the method remains useful beyond the original
classification task. These results are best interpreted as supporting
evidence for broader applicability, rather than as a complete forecasting study. The present work still has
important boundaries: the strongest evidence comes from one primary benchmark,
the shift analysis is concentrated on a specific leave-one-environment-out
weather protocol, and the objective weights are fixed rather than adapted during
training. Future work should therefore test broader types of shift, richer
multimodal settings, and adaptive balancing strategies for evidential and
consistency terms.

\section{Conclusion}
This paper introduced eCoMAE, an observability-aligned evidential masked pretraining method for multivariate time series. Rather than treating masked pretraining purely as a reconstruction problem, eCoMAE explicitly links masked modelling to uncertainty formation, so that the pretraining objective constrains not only what should be reconstructed, but also how confidence should behave under partial observability. To achieve this, eCoMAE combines block-wise time-step masking with a Normal--Inverse--Gamma evidential decoder and three complementary mechanisms: uncertainty-weighted consistency on overlap-visible positions, explicit evidence suppression on masked positions, and representation consistency across stochastic mask realisations.

The empirical results support this design in two ways. On IMPTC VRU intention prediction under environment shift, eCoMAE improves reliability-related performance, including calibration and selective-prediction behaviour, while maintaining competitive predictive accuracy. The ablations further indicate that these gains do not arise from stronger reconstruction or increased model capacity alone, but from coupling the evidential objective with observability-aligned consistency. Beyond the primary benchmark, additional evaluation on pedestrian trajectory forecasting across IMPTC, Waymo, nuScenes, and inD provides supporting evidence that the method remains useful beyond a single downstream task.

Overall, the results indicate that masked pretraining can be made more reliable under distribution shift when evidence formation is explicitly tied to observability. The present study is nevertheless bounded by the concentration of the deepest analysis on one primary benchmark and by the use of fixed objective weights. Future work will study broader forms of shift, richer multimodal temporal settings, and more adaptive balancing strategies for evidential and consistency terms.

\section*{Acknowledgments}
This research has been partly funded by the Hessian State Ministry for Higher Education, Research, Science and the Arts (HMWK) as part of the LOEWE program (Kapitel 15 02, F\"orderbuchungskreis 2995, F\"orderprodukt 11 -- Landes-Offensive zur Entwicklung Wissenschaftlich-\"okonomischer Exzellenz).

\paragraph{Use of Generative AI.}
Generative AI tools were used only for language polishing and readability improvement.
All technical content, experimental design, results, and claims were produced and verified by the authors.
\bibliographystyle{unsrt}
\bibliography{refs}

\clearpage
\appendix
\section{Detailed Ablations and Reproducibility Settings}
\label{app:ablations}
We provide the exact experiment settings and additional OOD uncertainty metrics underlying the full eight-metric ablation table in the main text. All 18 variants follow the same experiment order as Table~\ref{tab:main_18_variants_compact}; EVI denotes the evidential NIG objective and REP denotes cross-view representation consistency.

\subsection{Evidential-objective configuration dictionary}
Table~\ref{tab:evi_config_dictionary_correct} specifies the weights and cap values used by each ablation. Unless explicitly changed, the default configuration uses $\lambda_{\mathrm{evi}}=1$, $\lambda_{\mathrm{reg}}=10^{-2}$, $\bar\Phi=2$, $\lambda_{\mu}=\lambda_{\alpha}=0.05$, and $\lambda_{\mathrm{rep}}=0.10$. In the component ablations, the NIG objective and REP remain active while the designated consistency terms are selectively deactivated.

\begin{table}[H]
\centering
\footnotesize
\setlength{\tabcolsep}{3.2pt}
\renewcommand{\arraystretch}{1.14}
\begin{tabular*}{\linewidth}{@{\extracolsep{\fill}}lcccccc@{}}
\toprule
\textbf{Ablation setting} & $\lambda_{\mathrm{evi}}$ & $\lambda_{\mathrm{reg}}$ & $\bar\Phi$ & $\lambda_{\mu}$ & $\lambda_{\alpha}$ & $\lambda_{\mathrm{rep}}$ \\
\midrule
\rowcolor{gshade}\multicolumn{7}{@{}l@{}}{\textbf{Evidential objective (REP off)}} \\
MAE (no EVI/REP) & 0.0 & 0.0 & 0.0 & 0.00 & 0.00 & 0.00 \\
EVI (no REP) & 1.0 & $10^{-2}$ & 2.0 & 0.05 & 0.05 & 0.00 \\
\rowcolor{gshade}\multicolumn{7}{@{}l@{}}{\textbf{Representation consistency (EVI on)}} \\
$\lambda_{\mathrm{rep}}=0.05$ & 1.0 & $10^{-2}$ & 2.0 & 0.05 & 0.05 & 0.05 \\
eCoMAE ($\lambda_{\mathrm{rep}}=0.10$) & 1.0 & $10^{-2}$ & 2.0 & 0.05 & 0.05 & 0.10 \\
$\lambda_{\mathrm{rep}}=0.20$ & 1.0 & $10^{-2}$ & 2.0 & 0.05 & 0.05 & 0.20 \\
\rowcolor{gshade}\multicolumn{7}{@{}l@{}}{\textbf{Evidential consistency (EVI + REP on)}} \\
No $\mu/\alpha$ consistency & 1.0 & $10^{-2}$ & 2.0 & 0.00 & 0.00 & 0.10 \\
$\mu$ consistency only & 1.0 & $10^{-2}$ & 2.0 & 0.05 & 0.00 & 0.10 \\
$\alpha$ consistency only & 1.0 & $10^{-2}$ & 2.0 & 0.00 & 0.05 & 0.10 \\
2$\times$ $\mu/\alpha$ weights & 1.0 & $10^{-2}$ & 2.0 & 0.10 & 0.10 & 0.10 \\
\rowcolor{gshade}\multicolumn{7}{@{}l@{}}{\textbf{Evidential loss weight}} \\
$\lambda_{\mathrm{evi}}=0.1$ & 0.10 & $10^{-2}$ & 2.0 & 0.05 & 0.05 & 0.10 \\
$\lambda_{\mathrm{evi}}=5$ & 5.0 & $10^{-2}$ & 2.0 & 0.05 & 0.05 & 0.10 \\
\rowcolor{gshade}\multicolumn{7}{@{}l@{}}{\textbf{Regularisation and evidence cap}} \\
$\lambda_{\mathrm{reg}}=0$ & 1.0 & 0.0 & 2.0 & 0.05 & 0.05 & 0.10 \\
$\lambda_{\mathrm{reg}}=10^{-3}$ & 1.0 & $10^{-3}$ & 2.0 & 0.05 & 0.05 & 0.10 \\
$\lambda_{\mathrm{reg}}=0.1$ & 1.0 & $10^{-1}$ & 2.0 & 0.05 & 0.05 & 0.10 \\
$\bar{\Phi}=2.5$ & 1.0 & $10^{-2}$ & 2.5 & 0.05 & 0.05 & 0.10 \\
$\bar{\Phi}=6$ & 1.0 & $10^{-2}$ & 6 & 0.05 & 0.05 & 0.10 \\
\rowcolor{gshade}\multicolumn{7}{@{}l@{}}{\textbf{REP-only controls (EVI off)}} \\
$\lambda_{\mathrm{rep}}=0.10$ & 0.0 & 0.0 & 0.0 & 0.00 & 0.00 & 0.10 \\
$\lambda_{\mathrm{rep}}=0.20$ & 0.0 & 0.0 & 0.0 & 0.00 & 0.00 & 0.20 \\
\bottomrule
\end{tabular*}
\caption{Hyperparameters for the 18 experimental variants (same order as Table~\ref{tab:main_18_variants_compact}). Evidence-cap values ($\bar\Phi=2.5$ and $6$) follow the main-text ablation study.}
\label{tab:evi_config_dictionary_correct}
\end{table}

\clearpage
\subsection{OOD selective-prediction and uncertainty diagnostics}
Table~\ref{tab:ood_aurc_uecorr_compact} complements calibration results with the area under the risk--coverage curve (AURC) and uncertainty--error correlation (UE-corr), for both probing and finetuning. Lower AURC and higher UE-corr indicate stronger selective-prediction reliability and error awareness, respectively.

\begin{table}[H]
\centering
\small
\setlength{\tabcolsep}{5.2pt}
\renewcommand{\arraystretch}{1.16}
\begin{tabular*}{\linewidth}{@{\extracolsep{\fill}}lcccc@{}}
\toprule
\multirow{2}{*}{\textbf{Ablation setting}} & \multicolumn{2}{c}{\textbf{Probe}} & \multicolumn{2}{c}{\textbf{Finetune}} \\
\cmidrule(lr){2-3}\cmidrule(l){4-5}
& AURC$_p\downarrow$ & UE-corr$_p\uparrow$ & AURC$_f\downarrow$ & UE-corr$_f\uparrow$ \\
\midrule
\rowcolor{gshade}\multicolumn{5}{@{}l@{}}{\textbf{Evidential objective (REP off)}} \\
MAE (no EVI/REP) & 0.1807 & 0.2082 & 0.0681 & 0.4397 \\
EVI (no REP) & 0.1660 & 0.2615 & 0.0666 & 0.4538 \\
\rowcolor{gshade}\multicolumn{5}{@{}l@{}}{\textbf{Representation consistency (EVI on)}} \\
$\lambda_{\mathrm{rep}}=0.05$ & 0.1656 & 0.2616 & 0.0660 & \textbf{0.4740} \\
eCoMAE ($\lambda_{\mathrm{rep}}=0.10$) & 0.1658 & 0.2633 & 0.0656 & 0.4500 \\
$\lambda_{\mathrm{rep}}=0.20$ & 0.1651 & 0.2596 & 0.0680 & 0.4354 \\
\rowcolor{gshade}\multicolumn{5}{@{}l@{}}{\textbf{Evidential consistency (EVI + REP on)}} \\
No $\mu/\alpha$ consistency & 0.1673 & 0.2589 & \textbf{0.0646} & 0.4450 \\
$\mu$ consistency only & 0.1631 & 0.2585 & 0.0660 & 0.4380 \\
$\alpha$ consistency only & 0.1644 & 0.2628 & 0.0660 & 0.4638 \\
2$\times$ $\mu/\alpha$ weights & 0.1631 & 0.2655 & 0.0658 & 0.4575 \\
\rowcolor{gshade}\multicolumn{5}{@{}l@{}}{\textbf{Evidential loss weight}} \\
$\lambda_{\mathrm{evi}}=0.1$ & \textbf{0.1589} & 0.2743 & 0.0676 & 0.4627 \\
$\lambda_{\mathrm{evi}}=5$ & 0.1659 & \textbf{0.2856} & 0.0674 & 0.4722 \\
\rowcolor{gshade}\multicolumn{5}{@{}l@{}}{\textbf{Regularisation and evidence cap}} \\
$\lambda_{\mathrm{reg}}=0$ & 0.1647 & 0.2625 & 0.0658 & 0.4462 \\
$\lambda_{\mathrm{reg}}=10^{-3}$ & 0.1649 & 0.2625 & 0.0659 & 0.4615 \\
$\lambda_{\mathrm{reg}}=0.1$ & 0.1669 & 0.2598 & 0.0660 & 0.4712 \\
$\bar{\Phi}=2.5$ & 0.1639 & 0.2660 & 0.0682 & 0.4462 \\
$\bar{\Phi}=6$ & 0.1640 & 0.2578 & 0.0661 & 0.4404 \\
\rowcolor{gshade}\multicolumn{5}{@{}l@{}}{\textbf{REP-only controls (EVI off)}} \\
$\lambda_{\mathrm{rep}}=0.10$ & 0.1762 & 0.2141 & 0.0659 & 0.4664 \\
$\lambda_{\mathrm{rep}}=0.20$ & 0.1742 & 0.2514 & 0.0667 & 0.4480 \\
\bottomrule
\end{tabular*}
\caption{OOD uncertainty quality for the same 18 ablation variants. Lower AURC and higher UE-corr are better; best entries in each column are bold. Subscripts $p$ and $f$ indicate linear probing and finetuning, respectively.}
\label{tab:ood_aurc_uecorr_compact}
\end{table}
\FloatBarrier

\end{document}